\documentclass[letterpaper]{article} 
\usepackage[preprint]{aaai2027}  
\usepackage[hyphens]{url}  
\usepackage{graphicx} 
\usepackage{natbib}  
\usepackage{caption} 
\usepackage{algorithm}
\usepackage{algorithmic}
\usepackage{newfloat}
\usepackage{listings}
\DeclareCaptionStyle{ruled}{labelfont=normalfont,labelsep=colon,strut=off} 
\floatstyle{ruled}
\newfloat{listing}{tb}{lst}{}
\floatname{listing}{Listing}
\usepackage{booktabs}
\usepackage{tikz}
\usetikzlibrary{arrows.meta,positioning,calc}

\newcommand{\lllsym}{\mathbin{{<}\mkern-6mu{<}\mkern-6mu{<}}}
\newcommand{\lllop}{\lllsym}

\title{Long-Horizon State Tracking in LLMs: Executing MD5 through a Deep Sequence of Dependent Tool Calls}

\author{
    Dheeraj Mohandas Pai\equalcontrib,
    Lu Xian\equalcontrib
}
\affiliations{
    Leanmcp.com\\
    \{dheeraj.pai, lu.xian\}@leanmcp.com
}

\begin{document}

\maketitle

\begin{abstract}
Long-horizon tasks remain uncommon in large language model (LLM) evaluation, and
for a reason: when each step depends on the last, per-step accuracy that looks
excellent in isolation decays catastrophically, as errors cascade and the
end-to-end failure probability grows sharply with length. Existing agentic
benchmarks report end-to-end success but confound this state-tracking difficulty
with instruction interpretation, give no control group that isolates it, and are
vulnerable to shortcuts such as a hallucinated final answer, so they cannot say
\emph{why} a long run fails. Whether an LLM can carry exact intermediate state
across many tool calls at all is itself not well established. We test this cleanly
by having the model compute a cryptographic hash, MD5, \emph{step by step}: a sequence
of $196$ dependent tool calls over $64$ rounds while it carries four $32$-bit
words $(a,b,c,d)$ in its own context from one call to the next. Interpretation is
trivial and, because we implement MD5 from scratch (RFC~1321), we align every call
to the ground-truth trace and check the digest to the bit, so any failure is pure
bookkeeping. gpt-oss-120b, a mixture-of-experts model with only $\sim$5.5B active
parameters per token, at temperature $0$ with a short fixed prompt, carries the
full state across all $196$ calls and returns the correct digest on a majority of
completed runs. In the strongest setting we replace every primitive tool with a
second LLM, so a \emph{driver} and a \emph{worker} compute the whole hash from
scratch with no exact-arithmetic oracle in the loop. Two ingredients decide
success and neither changes the weights: keeping the model's own reasoning in its
context each turn (stripping it makes success collapse), and voting over a
thinking-enabled worker to remove its modular-arithmetic slips. We localize the
residual failures by origin, separating state-carrying from arithmetic and from
serving.
\end{abstract}

\section{Introduction}

\begin{figure}[tb]
\centering
\includegraphics[width=0.72\columnwidth]{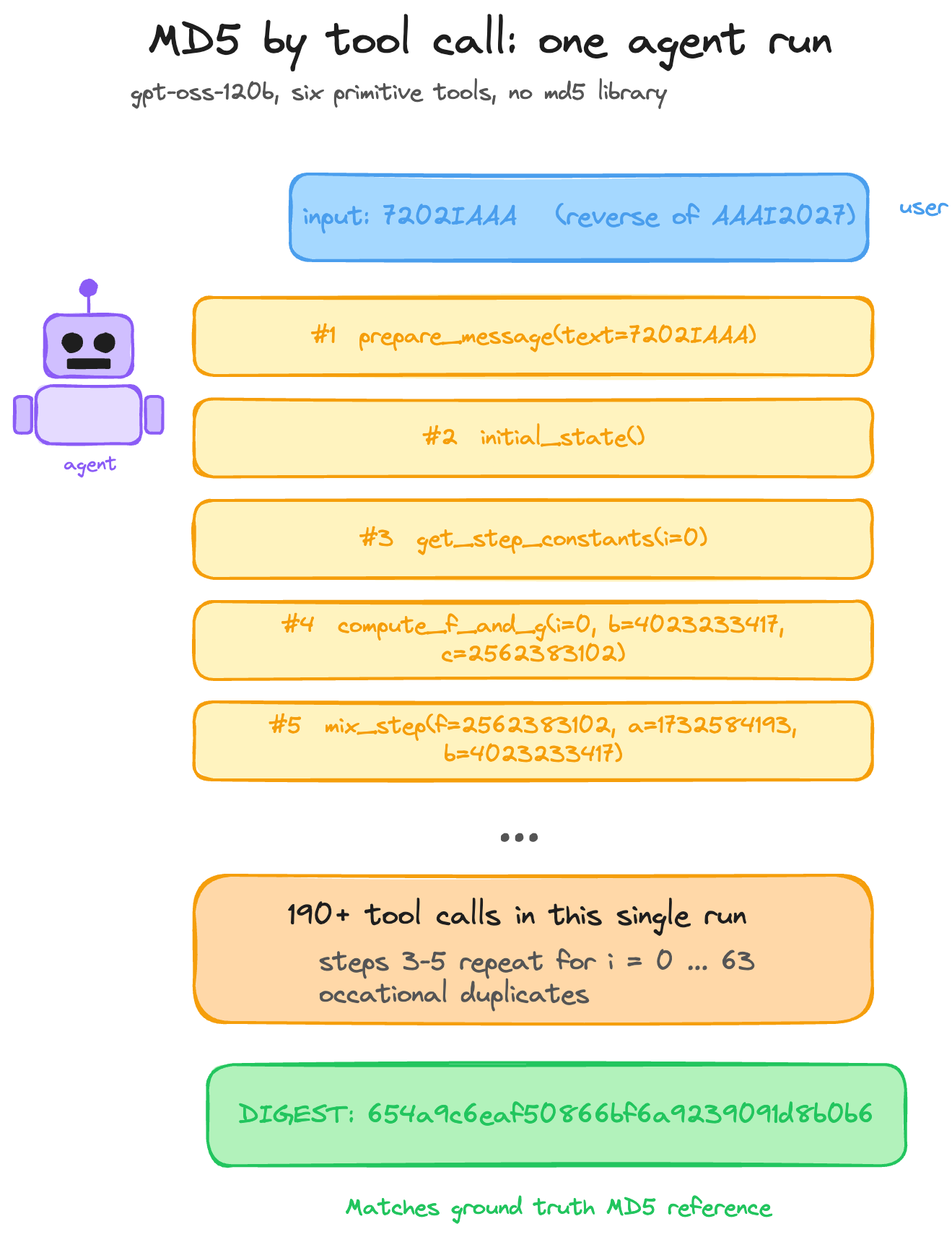}
\caption{A single agent run (driver with deterministic tools). The driver
(gpt-oss-120b) is given the input and the seven primitive tools and issues the
calls itself: \texttt{prepare\_message}, \texttt{initial\_state}, then
\texttt{get\_step\_constants} / \texttt{compute\_f\_and\_g} / \texttt{mix\_step}
repeated for rounds $i{=}0\ldots63$, carrying $(a,b,c,d)$ between calls. One hash
is $190$+ tool calls in a single run (with occasional benign duplicate lookups),
and the final digest matches the from-scratch MD5 reference.}
\label{fig:run}
\end{figure}

Long, deterministic, rule-governed procedures are widely regarded as a weak spot
for large language models (LLMs). The empirical case is strong. Single-step
accuracy that looks excellent in isolation decays sharply once a task requires
many \emph{dependent} steps, because per-step errors compound and models
``self-condition'' on their own earlier mistakes~\citep{sinha2025illusion}; and
quality degrades as the context grows past a few tens of thousands of tokens or a
few tens of turns~\citep{liu2024lostmiddle,modarressi2025nolima}. The usual
engineering response is to take the model out of the driver's seat and hand
control to a hand-authored controller, a decision tree, a workflow graph, a
finite-state machine, that owns the procedure while the LLM is confined to local
text tasks.

This paper asks whether that trade-off is actually forced, and answers: \emph{no.
Given the right context, an LLM can drive a long, exact, multi-step procedure to
completion by itself.} We make the claim under conditions chosen to be as
unforgiving as possible. Interpretation is trivial and every step is provably
right or wrong, so nothing can hide. The procedure is long and every step depends
on the last, so there is nowhere to get lucky. And in the strongest configuration
there is no exact-arithmetic oracle anywhere in the loop: even the primitive
computations are performed by an LLM.

Our testbed is \textbf{executing MD5 step by step}. MD5 over a
single $512$-bit block~\citep{rivest1992md5} is $64$ rounds maintaining four
$32$-bit words $(a,b,c,d)$. We expose each primitive (padding, per-round
constants, the round functions, the mixing step, the final addition) as a tool,
so the model's job is to call the tools in order and \emph{carry $(a,b,c,d)$ from
one call to the next}, a canonical sequence of $\mathbf{196}$ \textbf{dependent
tool calls}. This isolates exactly the capability in question: relaying exact
intermediate state across a long horizon, and taking many mechanical turns that a
computer finds trivial, \emph{consistently}. Because we implement MD5 from
scratch (no \texttt{hashlib}) as an exact reference, we can align the agent's
calls to the ground-truth trace and check the digest to the bit.

We then remove the safety net. In the strongest configuration the primitive tools
are not CPU code at all: each is replaced by a \textbf{worker LLM} that computes
the operation, $32$-bit bitwise $F/G/H/I$, modular addition, left rotation. The driver LLM and the worker LLM together compute the entire hash from
scratch, with a from-scratch CPU implementation used only to \emph{verify} values
for our analysis, never to supply them.

Driving gpt-oss-120b~\citep{gptoss2025}, a mixture-of-experts model with only
$\sim$5.5B \emph{active} parameters per token, at temperature $0$ with a short
fixed system prompt, we find it carries the full state across all $196$ calls and
reproduces the correct digest on non-memorized inputs on a majority of completed
runs. Success turns on two conditions, neither a change to the weights: keeping the
model's own reasoning in its context each turn, and using self-consistency
(majority voting) on the worker that computes the arithmetic. We analyze both, and
characterize the residual failures, in the body.

\paragraph{Contributions.}
\begin{enumerate}
\item A controlled, ground-truthed benchmark in which an LLM executes MD5 step by
step through tools, turning long-horizon exactness into a measurable,
bit-checkable task with a canonical $196$-call reference trace.
\item A \emph{positive} result: a $\sim$5.5B-active-parameter LLM executes the
full $196$-step procedure end-to-end and returns the correct digest, and, with a
worker LLM replacing every primitive, the driver/worker pair computes the hash
\emph{from scratch}, arithmetic included.
\item A \textbf{driver/worker} method with self-consistent workers that separates
state tracking (driver) from primitive arithmetic (worker), and shows the
arithmetic is made reliable by majority voting while the state is carried by the
driver.
\item An identification of the two context conditions that make it
work, reasoning-in-context and self-consistency, and a characterization of the
residual failures by origin (driver state, worker arithmetic, serving).
\end{enumerate}

\section{Background}

Two threads of prior work make our setting look, a priori, nearly impossible.

\paragraph{Error accumulation over horizon.} A high per-step accuracy $p$ yields
an end-to-end success rate of roughly $p^{n}$ over $n$ dependent steps; at
$n\approx 196$ even $p=0.99$ predicts a $\sim$14\% success rate, and $p=0.95$
predicts $\sim$$4\times 10^{-5}$. \citet{sinha2025illusion} formalize this: strong
single-step accuracy systematically overstates long-horizon capability, models
degrade as they condition on their own prior outputs, and larger models and
chain-of-thought~\citep{wei2022cot} push the horizon out but do not remove the
wall.

\paragraph{Long-context degradation.} Independently, retrieval and reasoning
quality drop as context length grows, the ``lost in the middle'' positional
effect~\citep{liu2024lostmiddle} and length-driven degradation~\citep{modarressi2025nolima},
with many models falling below half their short-context score by $\sim$32k
tokens, and function-calling accuracy degrading in long contexts
specifically~\citep{longfunceval}.

A $196$-step exact procedure sits squarely in the intersection of both hazards:
long horizon \emph{and} a growing transcript of tool calls and results. That is
exactly why success here is informative, and why the conditions that produce it
(reasoning-in-context, self-consistency, a driver/worker split) are the point.

MD5~\citep{rivest1992md5} maps a
message to a $128$-bit digest. The message is padded and split into $512$-bit
blocks; each block is read as sixteen $32$-bit words $M[0..15]$. A running state
of four $32$-bit words $(a,b,c,d)$ is initialized to fixed constants. Each block is
processed by $64$ rounds. Round $i$ computes a nonlinear function of three of the
state words, $f = F(b,c,d)$ where $F$ is bitwise selection $(b\wedge c)\vee(\neg
b\wedge d)$ for $i<16$ and the analogous $G,H,I$ for the later quarters, together
with a message index $g$. It then forms
\[
b \;\leftarrow\; b + \big((a + f + K[i] + M[g]) \lllop s[i]\big),
\]
where $K[i]$ is a per-round sine constant, $s[i]$ a per-round rotation amount,
$+$ is addition modulo $2^{32}$, and $\lllsym$ is a left rotation; finally the four
words are rotated $(a,b,c,d)\leftarrow(d,b',b,c)$. After $64$ rounds the block's
result is added, again modulo $2^{32}$, into the running state, and the final
$(a,b,c,d)$ is emitted little-endian as the digest. In our harness one block is a
canonical $196$ tool calls: two to set up, three per round (constants, function,
mix) across $64$ rounds, and two to finalize.

Two properties make this a demanding test. First, it is a
long, strictly dependent computation: the $196$ steps must each be exactly right,
and by design MD5 has no exploitable structure (its add-rotate-xor construction
targets the avalanche effect, so the digest cannot be guessed or interpolated,
only computed step by step). This makes it a clean probe of sustained exact
execution. Second, the per-step arithmetic, $32$-bit modular addition with carries
and bit rotations, is exactly the kind of large-number exact computation on which
LLMs are known to be brittle~\citep{lee2024arithmetic}, so the worker path stresses
a second, independent weakness.

\section{Related Work}

\paragraph{Long-horizon execution and long context.} Our result pushes against two
strands of pessimism. \citet{sinha2025illusion} show that high single-step
accuracy overstates multi-step capability because errors compound and models
self-condition on their own outputs, the analytical $p^{n}$ wall. Separately,
quality decays with context \emph{length}: the ``lost in the middle'' positional
effect~\citep{liu2024lostmiddle}, length-driven degradation past roughly $32$k
tokens~\citep{modarressi2025nolima}, and tool-calling specifically degrading in
long contexts~\citep{longfunceval}. A $196$-step transcript exercises both.

\paragraph{Reasoning traces and exact arithmetic.} That intermediate work must be
externalized to enable multi-step computation is the scratchpad
finding~\citep{nye2021scratchpad} and, at prompting time, chain-of-thought
\citep{wei2022cot}; our reasoning-channel result is its serving-time analogue.
The worker path stresses a second known weakness, exact arithmetic on large
numbers, where LLM competence is surface-form and magnitude
dependent~\citep{lee2024arithmetic}; we make it reliable with primitive-level
self-consistency~\citep{wang2023selfconsistency} rather than a new method.

\paragraph{Tool-using agents and self-managed memory.} Tool-augmented agents such
as ReAct~\citep{yao2023react} and Toolformer~\citep{schick2023toolformer} raise
capability but leave workflow state implicit in context, which is exactly what we
stress. Our forward direction, a model that edits its own context, connects to
self-editing memory architectures like MemGPT~\citep{packer2023memgpt}; our MD5
harness offers a bit-checkable testbed for whether such self-management preserves
exact state across a long horizon.

\section{Experimental Setup}

\paragraph{Reference.} We implement MD5 from scratch (RFC~1321; no
\texttt{hashlib}), validated against the RFC test vectors. This is ground truth:
for any input we can produce the exact digest and the exact canonical sequence of
primitive operations.

\begin{figure}[tb]
\centering
\includegraphics[width=0.92\columnwidth]{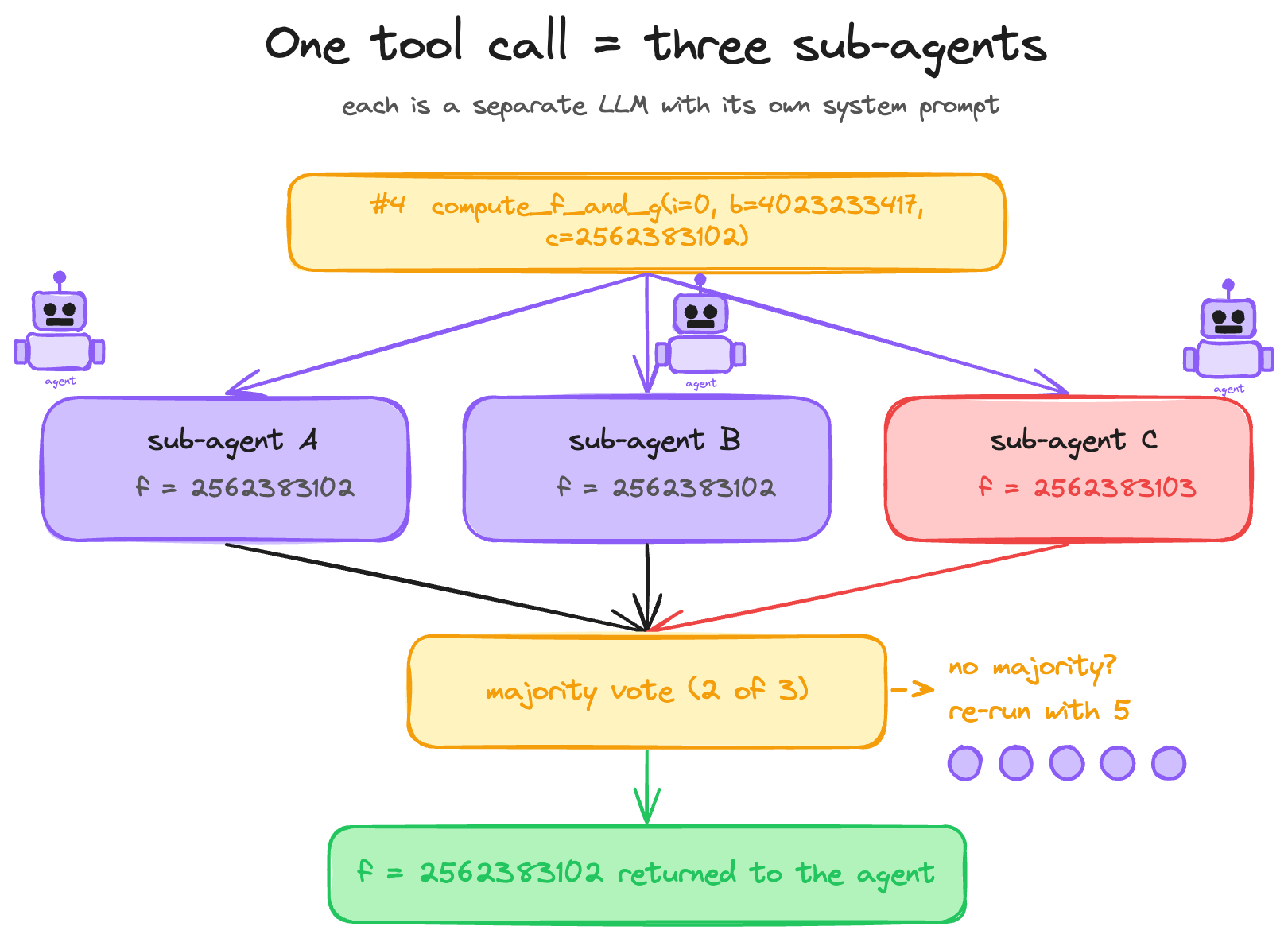}
\caption{One tool call computed by voting workers. In the strongest configuration
each primitive is a separate, stateless worker LLM. The operation is sampled three
times at temperature $0.7$ (each with its own fresh context) and the majority
value is returned to the driver; ties trigger an escalation round. This
self-consistency suppresses the worker's occasional modular-arithmetic slips, and
the worker's context is discarded after each call so only the driver stays
stateful.}
\label{fig:worker}
\end{figure}

\paragraph{Tools.} Seven primitives are exposed as tools:
\texttt{prepare\_message} (pad and split into sixteen $32$-bit words),
\texttt{initial\_state}, \texttt{get\_step\_constants($i$)},
\texttt{compute\_f\_and\_g($i,b,c,d$)} (round function $F/G/H/I$ plus the message
index), \texttt{mix\_step}, \texttt{add\_states}, and \texttt{to\_hex\_digest}.
Each is a thin exact wrapper over the reference (CPU mode) or an LLM twin (worker
mode). Figure~\ref{fig:run} shows a single driver run with deterministic tools,
and Figure~\ref{fig:worker} shows how a tool call is computed by voting workers.

\paragraph{Task.} A single-block input ($\le 55$ bytes, one $512$-bit block, $64$
rounds). The driver agent reads a \emph{short, fixed} instruction library and must
call the tools in order while carrying $(a,b,c,d)$ between calls and applying the
state-update rotation itself. A correct hash is a canonical $196$ tool calls ($2$
setup $+\,64\times 3\,+2$ finalize); the driver's final answer is a $32$-hex
digest compared for exact equality with $\mathrm{md5}(\mathrm{input})$.

\begin{algorithm}[tb]
\caption{MD5-by-hand driver loop (one $512$-bit block). Each \textsc{call} is a
tool the driver invokes; in swap mode a call is a worker-LLM majority vote, in CPU
mode a wrapper over the reference. The driver owns every variable below.}
\label{alg:driver}
\textbf{Input}: message \textit{text} (single block, $\le 55$ bytes)\\
\textbf{Output}: $32$-hex digest
\begin{algorithmic}[1]
\STATE $M \gets$ \textsc{prepare\_message}(\textit{text}) \quad// sixteen $32$-bit words
\STATE $(a,b,c,d) \gets$ \textsc{initial\_state}()
\STATE $(a_0,b_0,c_0,d_0) \gets (a,b,c,d)$
\FOR{$i = 0$ \TO $63$}
  \STATE $(K,s) \gets$ \textsc{get\_step\_constants}($i$)
  \STATE $(f,g) \gets$ \textsc{compute\_f\_and\_g}($i,b,c,d$)
  \STATE $\mathit{new\_b} \gets$ \textsc{mix\_step}($f,a,b,K,M[g],s$)
  \STATE $(a,d,c,b) \gets (d,c,b,\mathit{new\_b})$ \quad// rotation (most error-prone)
\ENDFOR
\STATE $(a,b,c,d) \gets$ \textsc{add\_states}($a_0,b_0,c_0,d_0,a,b,c,d$)
\STATE \textbf{return} \textsc{to\_hex\_digest}($a,b,c,d$)
\end{algorithmic}
\end{algorithm}

MD5 is a clean \emph{difficulty dial}:
interpretation is trivial and every primitive is provably correct, so what remains
to be tested is purely (i) whether the model can relay exact state between tools
without dropping any, and (ii) whether it can take many mechanical turns
\emph{consistently}. We fix a \emph{single block} deliberately, it keeps the
instruction library linear, and note that scaling the horizon is as easy as
adding blocks (each block is another $64$ rounds of the same loop), so the
benchmark extends to arbitrarily long horizons without adding interpretive
difficulty. We never use famous vectors (e.g.\ \texttt{"abc"}) whose digests may
be memorized; inputs are varied non-memorized single-block strings, including
random ones, so a correct digest cannot be recalled and must be computed.

\subsection{Models and configuration}

\paragraph{Models.} Driver and worker are both gpt-oss-120b, an open-weight
mixture-of-experts model with $\sim$117B total and $\sim$5.1--5.5B active
parameters per token, shipped with the MoE expert weights natively quantized to
MXFP4 ($\sim$4.25 bits), which is what all our endpoints serve.

\paragraph{Parameters.} The driver runs at temperature $0$ with a short fixed
system prompt and context caching enabled (the large invariant prefix, the
instructions and the constant tables, is cached so only the growing tail is
re-processed). In swap mode the worker runs at temperature $0.7$ with $3$-sample
majority voting. We repeat runs on multiple non-memorized single-block inputs.

\paragraph{Serving.} One provider serves the serverless and dedicated gpt-oss
endpoints; a second serves the fast, thinking-enabled worker with its own
low-precision kernels. We pin dates and log the endpoint per run; ``temperature
$0$'' is not bit-deterministic on quantized serving, so we treat every cell as a
distribution and report $N$.

\subsection{Verification and metrics}

\paragraph{Ground-truth alignment.} We build the ground-truth call sequence for
any input and align it call-for-call against what the agent actually did,
rendering a green/red trace, green where the agent's tool \emph{inputs} match
ground truth, red where they diverge. We inject each first divergence into the
real algorithm to produce a counterfactual digest (``if only this one slip had
occurred, the hash would be $X$''), separating a \emph{fresh} error from its
downstream propagation. A live mode flags the first state slip in real time,
keyed by round index so benign duplicate lookups do not misalign the comparison.

\paragraph{Worker verification.} When a primitive is computed by a
worker LLM, each swapped op is checked live against its CPU twin. This
verification is \emph{for measurement and human legibility only}, it is not a
corrective oracle in the reported ``from scratch'' runs. A policy knob can either
let a wrong worker value cascade (to observe end-to-end damage) or correct it in
place (to attribute a failure), and we report both.

\paragraph{Metrics.} \emph{Success rate} (primary): final digest equals true
$\mathrm{md5}(\mathrm{input})$, exact, reported as \emph{successes / runs}.
\emph{First-divergence round}: where the carried state first departs ground truth,
as a distribution over runs. \emph{Calls-to-completion} vs.\ the canonical $196$
(skips / duplicates / early stop). \emph{Per-op worker pass rate}: worker vs.\ CPU
twin, per primitive, with and without $3$-sample voting. \emph{Latency} (context
only). Every request, response, tool call, result, and worker sample (including
full reasoning) is logged as structured JSONL, indexed into SQLite with a
per-session success / wrong / no-digest label computed against $\mathrm{md5}()$.

\section{Results}

\subsection{State-carrying driver}

The seven primitives run as exact CPU tools; the driver carries $(a,b,c,d)$
between them. The driver emits the full canonical $196$-call sequence and
returns the correct $32$-hex digest, verified as a match, on multiple distinct
inputs: a $\sim$5.5B-active-parameter MoE model carries exact $32$-bit state from
step $0$ to step $196$. Against the $p^{n}$ prior, the horizon wall is not
fundamental once the context is managed. Serving, not weights, then governs the
rest. For the same weights, correctness and speed differ sharply by route
(Table~\ref{tab:feasibility}): a fast reasoning endpoint that is excellent
single-shot is a poor multi-turn driver, looping or stopping early and rarely
emitting a digest at all, which is itself evidence that the difficulty is
sustained state across turns rather than any single step. When a run does fail,
the tools are never wrong: the green/red trace goes red exactly where the driver
feeds a tool a mis-carried $32$-bit word and stays red as the corruption
propagates. The most common fresh error is the state-update rotation
($a,d,c,b \leftarrow d,c,b,\mathrm{new\_}b$); failed runs also show
dropped/duplicated calls and early stops (e.g.\ finalizing in $172$ rather than
$196$ calls).

\begin{table}[tb]
\centering
\setlength{\tabcolsep}{1.1mm}
\begin{tabular}{lrrrr}
\toprule
Driver endpoint & Succ. & Wrong & Compl. & Rate \\
(gpt-oss-120b, echo on) & & & & \\
\midrule
Groq       & $0$  & $2$  & $2$  & $0\%$  \\
Fireworks  & $21$ & $13$ & $34$ & $62\%$ \\
\bottomrule
\end{tabular}
\caption{Feasibility of the state-carrying driver, current run corpus. ``Compl.''
counts runs that produced a $32$-hex digest (success or wrong); the $71$ Fireworks
runs and $71$ Groq runs that never emitted a digest (loops / early stops / aborted
development runs) are excluded from the ratio. The Fireworks row pools all of that
provider's gpt-oss-120b endpoints. Cells are raw counts from the logged corpus, to
be finalized on a frozen $N\ge 20$ grid.}
\label{tab:feasibility}
\end{table}

\subsection{Removing the arithmetic oracle}

\paragraph{Design.} Every primitive is swapped onto a \textbf{worker LLM}, also
gpt-oss-120b, but served on a fast, thinking-enabled endpoint as a single-shot,
stateless computer. The worker runs at temperature $0.7$ with $3$-sample majority
voting; the majority value is returned to the driver, which continues its
$196$-call journey. The driver stays at temperature $0$. The CPU twin records a
pass/fail for each op for our analysis but does not supply values.

\paragraph{Findings.} With majority voting, per-op worker pass rates are high
across all six computed primitives, so the driver/worker pair computes the hash
from scratch with no exact oracle in the loop. The characteristic worker error is
in the modular reduction of \texttt{mix\_step}/\texttt{add\_states}: it reasons its
way to the correct unreduced sum but subtracts the wrong multiple of $2^{32}$
(e.g.\ $-1\times$ the modulus where it needs $-2\times$). The worker frequently
catches and fixes this itself mid-reasoning by comparing magnitudes, and what it
misses in one sample is corrected by the majority vote across three. When a full
run does fail under swap-all it is overwhelmingly the driver mis-carrying state,
not the worker computing a primitive wrong: correcting every worker op in place
(removing arithmetic error entirely) does not by itself make a driver-slipped run
succeed, so the two error sources are independent and state tracking is the harder
one.

\subsection{What makes it work}

\paragraph{Reasoning in context.} The single
largest lever is whether the model's own reasoning trace is kept in its context
each turn. gpt-oss emits its intermediate work in a dedicated reasoning channel
(the ``analysis'' channel of OpenAI's Harmony format~\citep{openai2025harmony});
the question is whether that channel is fed back on the next turn. With it, the
driver tracks state across $196$ calls; stripped, success collapses on the
identical model, prompt, and input. Crucially, the failure is \emph{not} fixed by
relocating the same text: when we place the reasoning content in the visible
\emph{assistant} message instead of the Harmony reasoning channel, keeping the
tokens byte-for-byte identical, the driver still fails, and does so early, usually
within the first $12$ to $24$ of $196$ steps, looping or finalizing prematurely.
The effect is therefore specific to the reasoning channel being present as the
model's working memory, not to the tokens being visible somewhere. This mirrors
the scratchpad result~\citep{nye2021scratchpad} that multi-step computation
depends on where intermediate work lives, not merely that it exists. It is also
provider plumbing, not weights: some stacks keep the reasoning field on input,
others reject it (returning an HTTP $400$ on the echoed field), so the \emph{same
weights} behave differently by route. Consistent with this, most of the runs that
never produce a digest are early collapses: of the $105$ logged Fireworks
gpt-oss-120b sessions, $71$ emit no digest at all, and nearly half of those halt
within the first $24$ of $196$ calls. It is a clean ablation with a large success
delta and, we argue, the most actionable finding: manage the reasoning channel and
you get the horizon.

With high thinking effort the driver's state-carrying is reliable and the
worker's arithmetic, including modular reduction, is almost always right; the
remaining failures are knocked down further by increasing worker samples. The two
knobs that matter are: keep the thinking, and vote.

\paragraph{The $40\rightarrow 48$ skip.} A
striking failure recurs at a fixed location: among the runs that get past round
$47$, several skip exactly rounds $40$ through $47$, so the driver's per-round
tool calls jump straight from round $39$ to round $48$ and it finalizes with a
short block. In our corpus this appears in $7$ long runs and, tellingly, in
\emph{none} of the successful ones. It is contiguous and position-locked rather
than random: rounds $40$ to $47$ are the back half of the third ($H$) quarter, and
a couple of runs skip rounds $24$ to $31$ as well (the back half of the $G$
quarter), so the model appears to ``round off'' the second half of a $16$-round
block. We read this as a candidate model fingerprint rather than a claim that it
happens on every run.

For the same serverless
weights, route changes latency by $\sim$5$\times$ (a gateway path $\approx$2.5\,s
per step vs.\ a direct path $\approx$13\,s per step) and, historically, correctness
(dedicated vs.\ shared; quantization / precision). We therefore control the
endpoint per experimental cell and report speed only as context.

\subsection{Failure analysis}

Because the reference is exact, every failure has a precise location and cause.
Table~\ref{tab:failures} is the consolidated taxonomy; the counterfactual
injection (Sec.~Verification) lets us confirm that each failed run has a
\emph{single} originating error whose corruption then propagates. Three points are
worth drawing out. First, the errors partition cleanly by \emph{origin}: driver
(state), worker (arithmetic), or serving (context), and these are independent, so
removing one class (e.g.\ voting out worker slips) does not remove another.
Second, the driver errors dominate: the rotation slip and the dropped/duplicated
step account for most failed runs, consistent with state tracking, not arithmetic,
being the bottleneck. Third, one driver error is not random in \emph{where} it
lands. The $40\rightarrow 48$ skip recurs at the same location (and only in failing
runs), a position-locked fingerprint rather than a stochastic slip, and a
candidate diagnostic for the model family. By contrast the worker's modular
wrap-around is stochastic
and self-limiting, the worker often repairs it mid-reasoning by comparing
magnitudes, and majority voting removes almost all of what remains.

\begin{table}[tb]
\centering
\setlength{\tabcolsep}{1mm}
\begin{tabular}{@{}llp{3.2cm}@{}}
\toprule
Failure mode & Origin & Mechanism / effect \\
\midrule
State slip (rotation) & Driver & mis-applies $a,d,c,b\!\leftarrow\!d,c,b,\mathit{new\_b}$; wrong from that round on \\
Dropped / duplicated step & Driver & skips or repeats a call; trace misaligns \\
Early stop / no digest & Driver & finalizes early (e.g.\ $172$ vs $196$ calls) or loops without a digest \\
$40\!\rightarrow\!48$ skip \emph{(signature)} & Driver & jumps rounds $40$--$47$ at a fixed location (failing runs only); recurring fingerprint \\
Modular wrap-around & Worker & subtracts wrong multiple of $2^{32}$; one bad primitive, caught by vote / twin \\
Reasoning stripped & Serving & analysis channel not echoed; collapse within $12$--$24$ steps \\
\bottomrule
\end{tabular}
\caption{Failure taxonomy by origin. Driver (state-carrying) errors dominate;
worker (arithmetic) errors are rare and self-consistency-correctable; the serving
error is an all-or-nothing collapse. The $40\rightarrow 48$ skip is recurring and
position-locked (failing runs only), a candidate model signature rather than a
random slip.}
\label{tab:failures}
\end{table}

\section{Discussion}

\paragraph{Practical reach.} Long-horizon \emph{exactness} is not a toy concern.
Accounting and bookkeeping close-out, legal and compliance workflows, provisioning
and reconciliation pipelines, all require an intermediate state to be preserved
verbatim across many steps, where a single dropped or mis-copied value silently
corrupts the outcome. The prevailing assumption is that such workflows must be
driven by a hand-built controller with the LLM confined to leaf tasks. Our result
says the LLM can hold the state itself when its context is managed for
it, reasoning kept in context, arithmetic made reliable by voting, which widens
where a model may be trusted to \emph{drive}, not just assist.

\paragraph{Toward a self-managing agent.} The driver/worker
split is deliberately a two-agent decomposition: one agent owns the long-horizon
state and sequencing, the other owns bounded, verifiable computation. We use the
\emph{same} model for both roles on purpose. The split is scaffolding, not the
destination: because driver and worker are one model, the natural next step is to
collapse them into a single agent that carries its own state, computes its own
primitives, and, critically, curates its own context.

\paragraph{Self-editing context.} The most direct
version of that next step is to give the model a tool that edits its own context:
operations to write a value into a durable scratch region, read it back, summarize
or compress the transcript so far, and drop material it no longer needs. In our
task the state to be managed is tiny and exact, the four working words and the
round index, so a self-editing policy has an unusually clean success criterion:
we can check to the bit whether the model preserved the right state after every
edit, rather than judging memory quality qualitatively. This connects to
self-editing memory architectures such as MemGPT~\citep{packer2023memgpt}, which
expose paging between in-context and external memory as tool calls; our
contribution would be a ground-truthed testbed for whether such self-management
actually keeps exact state across a long horizon. Would this be useful? We expect
so: a model that can offload $(a,b,c,d)$ to a scratch tool and reload it on demand
should be far more robust to the length-driven degradation of long transcripts,
and the same mechanism generalizes to the branching, real-world procedures
(accounting close-outs, compliance workflows) where the state that must survive is
larger and messier than four words. Whether a single model can own its state
\emph{and} its computation \emph{and} its context is the question this setup is
designed to open.

\section{Conclusion}

Executing a deterministic algorithm step by step is a clean, bit-checkable probe of
long-horizon reliability, and it delivers a positive answer: given the right
context, a mixture-of-experts LLM with only $\sim$5.5B active parameters computes
an MD5 digest across $196$ dependent tool calls, and, with a self-consistent
worker replacing every primitive, does the whole thing, arithmetic included, from
scratch. Success turns on two levers that are about \emph{context}, not weights:
keep the model's reasoning in its context, and make the arithmetic reliable by
voting. The residual failures are few, mechanical, and, in the
$40\rightarrow 48$ skip, a reproducible signature of the model. The broader
lesson runs opposite to the usual prescription: rather than assume long, exact
workflows must be pulled out of the model and handed to a hand-built controller,
give the model the context it needs and it can drive them, a foundation for
agents that eventually manage that context themselves.

\section*{Acknowledgments}

We thank Google Cloud for research grant support, which provided the GPU access
used to run the experiments reported in this paper.

\bibliography{aaai2027}

\end{document}